\documentclass{article}

\usepackage{PRIMEarxiv}

\usepackage[utf8]{inputenc} % allow utf-8 input
\usepackage[T1]{fontenc}    % use 8-bit T1 fonts
\usepackage{hyperref}       % hyperlinks
\usepackage{url}            % simple URL typesetting
\usepackage{booktabs}       % professional-quality tables
\usepackage{amsfonts}       % blackboard math symbols
\usepackage{nicefrac}       % compact symbols for 1/2, etc.
\usepackage{microtype}      % microtypography
\usepackage{lipsum}
\usepackage{fancyhdr}       % header
\usepackage{graphicx}       % graphics
\graphicspath{{media/}}     % organize your images and other figures under media/ folder

\usepackage{amsmath,amsfonts,amscd,amssymb,bm}
\usepackage{graphicx}
\usepackage{epstopdf}
\usepackage{overpic}
\usepackage{cancel}
\usepackage{rotating}
\usepackage{url}
\usepackage{caption}
\usepackage{color}
\usepackage{rotating}
\usepackage{multirow}
\usepackage{wrapfig}
\usepackage{mathtools}
\usepackage{subeqnarray}
\usepackage{setspace}
\usepackage{subcaption}
\usepackage{soul, xcolor}
\usepackage[numbers,sort&compress]{natbib}
\usepackage{algorithm}
\usepackage{algpseudocode}
\usepackage{mathtools}
  
\title{Parametric PDE Control with Deep Reinforcement Learning and Differentiable L$_0$-Sparse Polynomial Policies}

\author{
  Nicolò Botteghi \\
  Department of Applied Mathematics \\
  University of Twente \\
  Enschede, The Netherlands\\
  \texttt{n.botteghi@utwente.nl} \\
   \And
  Urban Fasel \\
  Department of Aeronautics \\
  Imperial College London \\
  London, United Kingdom\\
  \texttt{u.fasel@imperial.ac.uk} \\
}

\begin{document}
\maketitle

\vspace{-.2in}
\begin{abstract}
Optimal control of parametric partial differential equations (PDEs) is crucial in many applications in engineering and science, such as robotics, aeronautics, chemisty, and biomedicine. In recent years, the progress in scientific machine learning has opened up new frontiers for the control of parametric PDEs. In particular, deep reinforcement learning (DRL) has the potential to solve high-dimensional and complex control problems in a large variety of applications. Most DRL methods rely on deep neural network (DNN) control policies. However, for many dynamical systems, DNN-based control policies tend to be over-parametrized, which means they need large amounts of training data, show limited robustness, and lack interpretability. In this work, we leverage dictionary learning and differentiable L$_0$ regularization to learn sparse, robust, and interpretable control policies for parametric PDEs. Our sparse policy architecture is agnostic to the DRL method and can be used in different policy-gradient and actor-critic DRL algorithms without changing their policy-optimization procedure. We test our approach on the challenging task of controlling a parametric Kuramoto-Sivashinsky PDE. We show that our method (1) outperforms baseline DNN-based DRL policies, (2) allows for the derivation of interpretable equations of the learned optimal control laws, and (3) generalizes to unseen parameters of the PDE without retraining the policies. 

\vspace{4pt}
\noindent Code available on \href{https://github.com/nicob15/Parametric-PDE-Control-with-DRL-and-Differentiable-L0-Sparse-Polynomial-Policies}{github.com/nicob15/PDE-Control-with-DRL} and \href{https://github.com/nicob15/Sparsifying-Parametric-Models-with-L0}{github.com/nicob15/Sparse-Parametric-Models}.
\end{abstract}

\noindent\emph{Deep Reinforcement Learning, Parametric Partial Differential Equations, Sparse Dictionary Learning, L$_0$ Regularization}

\section{Introduction}
Many systems in engineering and science can be modeled by partial differential equations (PDEs). The control of these systems governed by PDEs poses a complex computational challenge \cite{manzoni2021optimal}. Optimal control theory has been extensively studied \cite{lewis2012optimal} and has been used to solve different PDE control problems in a wide range of applications~\cite{ krstic2008boundary, troltzsch2010optimal,Kaiser2018prsa}, from robotics and engineering \cite{stengel1994optimal, goza2017strongly, hickner2023data} to chemistry and biomedicine \cite{miranda2008integrating}. Optimal control requires a (forward) dynamics model of the system. These models are traditionally derived from first principles or identified directly from data, e.g. by using system identification techniques \cite{ljung1998system}. Solving optimal control problems is computationally expensive due to the underlying optimization process that repeatedly needs to integrate the model forward in time, which is especially expensive for high-dimensional systems (\textit{curse of dimensionality}). Additionally, optimal control strategies are often sensitive to changes in the system parameters and require solving the optimal control problem for each new instance of the parameters. This makes developing PDE control techniques that are computationally efficient and adaptable to changes in the system parameters very challenging.

The recent progress in scientific machine learning has drastically changed how we can discover and analyze dynamical systems. 
This progress has permeated into the field of control under the name of reinforcement learning (RL) \cite{sutton2018reinforcement}. RL aims to solve control problems (i.e., sequential decision-making processes in the RL jargon), by learning an optimal control law (the policy), while interacting with a dynamical system (the environment). RL assumes no prior knowledge of the dynamics of the system, making RL a broadly applicable control approach when deriving or learning a dynamical system model is difficult. Deep reinforcement learning (DRL) is the extension of RL using \textit{deep} neural network (DNN) policies~\cite{arulkumaran2017deep, li2017deep, franccois2018introduction}. 
DRL has shown outstanding capabilities in controlling complex and high-dimensional systems such as games \cite{mnih2015human, ye2020mastering, shao2019survey, lample2017playing, mnih2013playing, van2016deep, wang2016dueling}, simulated and real-world robotics \cite{lin1992reinforcement, kober2013reinforcement, polydoros2017survey, botteghi2020reward, zhang2015towards, gu2017deep, zhao2020sim, botteghi2021low}, and recently PDEs \cite{bae2022scientific, beintema2020controlling, buzzicotti2020optimal, rabault2020deep, vignon2023recent, fan2020reinforcement, rabault2019accelerating, xia2023active, peitz2023distributed, zolman2024sindy}.

While DNNs are critical for the success of DRL, there are several drawbacks when using them, namely (i) their sample inefficiency and high training cost, (ii) limited robustness and generalization, (iii) and lack of interpretability of the learned policies. In particular, the DNN architectures used in DRL usually contain millions of parameters and need large amounts of training data and computations, which makes them unsuitable in most real-world applications where collecting data is expensive and time-consuming. Also, the control policies are often brittle to changes in the system parameters and fail when applied in different contexts or tasks without retraining. Additionally, the learned DNN policies are difficult to interpret and explain, limiting the possibility to perform stability or robustness analysis. These limitations are especially severe when dealing with scientific and engineering applications governed by PDEs due to the high computational cost of numerically solving PDEs, the possible variation of parameters of the PDE systems, and the missing interpretability and a posteriori analyses needed for deploying such DNN-based controllers on safety-critical real-world systems.  These challenges have been the core focus of recent research in the field, and many methods have been proposed to improve the sample efficiency, generalization, and interpretability of DRL. 
Successful methods include unsupervised representation learning \cite{lesort2018, botteghi2022unsupervised}, i.e. the problem of learning reduced-order representation of the data; imitation learning and behavioral cloning \cite{hussein2018, torabi2018behavioral, codevilla2019exploring}, i.e. the use of expert trajectories to initialize the policies; transfer learning \cite{torrey2010transfer, weiss2016survey}, i.e. the process of transferring learned strategies to new situations; meta-learning \cite{hospedales2021meta, vilalta2002perspective}, i.e. the concept of learning policies that can quickly adapt to new scenarios; and explainable artificial intelligence \cite{dovsilovic2018explainable}, i.e. tools to help understand and interpret DNN predictions. 

\begin{figure}[h!]
    \centering
    \includegraphics[width=1.0\textwidth, page=1]{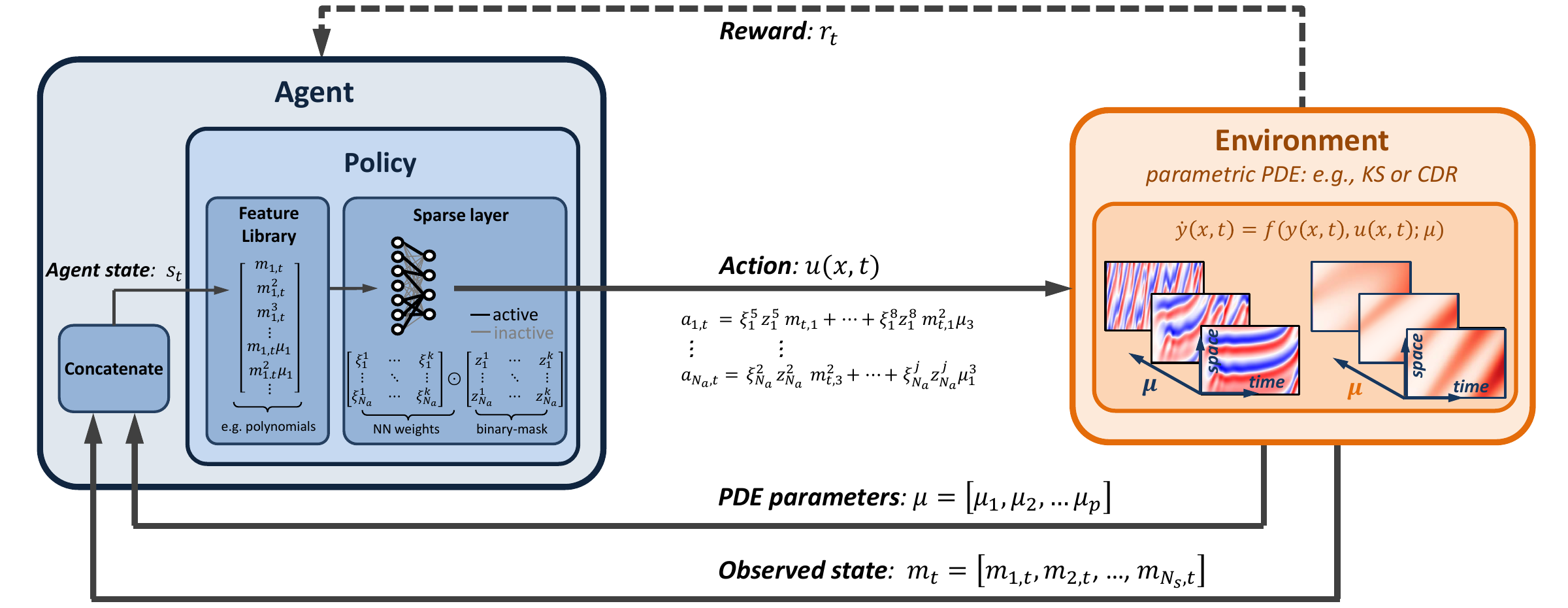}
    \caption{Deep reinforcement learning with L$_0$-sparse polynomial policies.}
    \label{fig:framework}
\end{figure}
In our work, we combine DRL with ideas from dictionary learning and differentiable L$_0$ regularization. 
We introduce our method starting from the observation that, for many dynamical systems, optimal control policies using DNNs tend to be over-parametrized. 
Sparse and simple (e.g. single-layer) NNs may be sufficiently expressive to approximate optimal control laws, and are more data efficient, quicker to train, and more interpretable compared to dense and complex DNNs \cite{louizos2018learning}. 
Moreover, sparsifying or pruning DNNs to reduce the number of parameters not only reduces the training cost, but also improves the prediction robustness of the DNN \cite{han2015deep, zhang2021understanding, molchanov2017variational, ullrich2017soft, louizos2018learning}. 
Sparsity can be enforced using the L$_0$ norm, which is non-convex and non-continuous, and therefore non-differentiable and difficult to introduce in gradient-based optimization used in DRL. In our work, we use pruning techniques in combination with dictionary learning \cite{mairal2009online, brunton2016discovering} to identify sparse DRL policies with a limited number of trainable parameters for the control of parametric PDEs. The method is shown in Figure \ref{fig:framework}. A policy function is learned, mapping the observed PDE states $m$ and the parameters $\mu$ to control inputs $u$, by a sparse linear combination of candidate nonlinear functions from a chosen dictionary, i.e., polynomials (although not limited to polynomials).
The polynomial features are fed to a single-layer neural network, with parameters corresponding to the coefficients of the polynomial, that are sparsified using a differentiable version of the L$_0$ norm \cite{louizos2018learning}. This implementation of the L$_0$ norm is suitable for gradient-based optimization techniques, e.g. stochastic gradient descent (SGD) or ADAM \cite{kingma2014adam}, and is compatible with any DRL algorithm. Therefore, our sparse polynomial policy can be directly and simply used to replace the DNN-based policies in any policy-based and actor-critic DRL algorithm. Additionally, the choice of the library allows us to not only avoid the use of a DNN but also to inject any available prior knowledge into the learning process. 

The paper is organized as follows: in Section \ref{sec:related_work} we introduce related work, and in Section \ref{sec:preliminaries} we introduce the building blocks of our framework, namely RL, sparse dictionary learning, and L$_0$ regularization. In Section \ref{sec:method}, we describe our L$_0$ sparse dictionary control policies method, and in Section \ref{sec:results} and \ref{sec:discussion} we show the results and discuss the findings.

\section{Related Work}\label{sec:related_work}
DRL has shown outstanding capabilities in controlling complex and high-dimensional systems such as games \cite{mnih2015human, ye2020mastering, shao2019survey, lample2017playing, mnih2013playing, van2016deep, wang2016dueling}, simulated and real-world robotics \cite{lin1992reinforcement, kober2013reinforcement, polydoros2017survey, botteghi2020reward, zhang2015towards, gu2017deep, zhao2020sim, botteghi2021low}, and recently dynamical systems and PDEs \cite{bae2022scientific, beintema2020controlling, buzzicotti2020optimal, rabault2020deep, vignon2023recent, fan2020reinforcement, rabault2019accelerating, xia2023active, peitz2023distributed, zolman2024sindy}. However, this success often came at the price of long training times, large amounts of data, high computational costs, and limited interpretability of the learned controllers. 

To improve upon the aforementioned limitations, several ideas from machine learning have been applied to DRL, such as unsupervised representation learning \cite{lesort2018, botteghi2022unsupervised}, i.e., the problem of learning reduced-order representation of the data; imitation learning and behavioral cloning \cite{hussein2018, torabi2018behavioral, codevilla2019exploring}, i.e., the use of expert trajectories to initialize the policies; transfer learning \cite{torrey2010transfer, weiss2016survey}, i.e., the process of transferring learned strategies to new situations; meta-learning \cite{hospedales2021meta, vilalta2002perspective}, i.e., the concept of learning policies that can quickly adapt to new scenarios; explainable artificial intelligence \cite{dovsilovic2018explainable}, i.e., tools to help understand and interpret DNN predictions; and sparsifying or pruning DNNs, i.e., to reduce the number of parameters not only reduces the training cost, but also to improve the prediction robustness of the DNN \cite{han2015deep, zhang2021understanding, molchanov2017variational, ullrich2017soft, louizos2018learning}. 

In the context of dynamical systems and control, sparse dictionary learning has been used to discover governing laws and build accurate models of the systems dynamics, yet with limited number of learnable parameters, that can be used for forecasting and control \cite{brunton2016sparse, fasel2022ensemble}. Sparse dictionary learning methods make typically use of the L$_1$ norm as a proxi for L$_0$ norm to find sparse solution. However, in our work we exploit the differentiable L$_0$ formulation in \cite{louizos2018learning} to learn control policies with limited number of active coefficients.

\section{Preliminaries}\label{sec:preliminaries}

\subsection{Reinforcement Learning}\label{subsec:reinforcement_learning}

Reinforcement learning (RL) is learning how to map situations to actions to maximize a numerical reward signal \cite{sutton2018reinforcement}.
RL has its foundation in the theory of Markov decision processes (MDP). An MDP is a tuple $\langle \mathcal{S}, \mathcal{A}, T, R \rangle$ where $\mathcal{S}  \subset \mathbb{R}^{N_s}$ is the set of observable states\footnote{It is worth noticing that the state in RL does not need to necessarily coincide with the full state of the PDE. Therefore, we utilize two different symbols to prevent confusion. The same holds for the control inputs.}, $\mathcal{A} \subset \mathbb{R}^{N_a}$ is the set of actions, $T: \mathcal{S} \times \mathcal{S} \times \mathcal{A} \longrightarrow [0,1]; \ (s',s,a) \longmapsto T(s',s,a)=p(s'|s,a)$ is the transition function, describing the probability $ps'|s,a)$ of reaching state $s'$ from state $s$ while taking action $a$, and $R: \mathcal{S} \times \mathcal{A} \longrightarrow \mathbb{R}; \ (s,a)\longmapsto R(s,a)$ is the reward function\footnote{The reward function is the analogous of the cost function in optimal control but with the opposite sign.}.
Any RL algorithm aims to find the optimal policy $\pi$, i.e. the mapping of states to actions, maximizing the total cumulative return~$G_t$:
\begin{equation}
G_t = \mathbb{E}_{\pi}[r_0 + \gamma r_1  + \gamma^2 r_2 + \cdots + \gamma^t r_t + \cdots ] = \mathbb{E}_{\pi}[\Sigma_{t=0}^{H} \gamma^t r_t],
\label{eq:cumulative_rew}
\end{equation}
where the subscript $t$ indicates the timestep, $\gamma$ is the discount factor, $r_t$ is the instantaneous reward received by the agent at timestep $t$, and $H$ the control horizon that can be either finite or infinite\footnote{In this work, we consider control problems with a finite horizon $H$.}.
An important element of RL is the notion of value function. The value function $V:\mathcal{S}\rightarrow\mathbb{R}$ and action-value function $Q:\mathcal{S} \times \mathcal{A} \rightarrow \mathbb{R}$ quantify the total cumulative return $G_t$ (Equation \eqref{eq:cumulative_rew}) of a state $s$ or state-action pair $(s, a)$. 

RL algorithms are usually classified as value-based, policy-based, or actor-critic \cite{sutton2018reinforcement}. 
In particular, actor-critic algorithms learn value function and policy at the same time. Actor refers to the policy acting on the environment, and critic refers to the value function assessing the quality of the policy. Examples are deep deterministic policy gradient (DDPG) \cite{lillicrap2015continuous}, proximal policy optimization (PPO) \cite{schulman2017proximal}, and soft actor-critic (SAC) \cite{haarnoja2018soft}.

In this work, we present a method that can replace any policy function of policy-based and actor-critic methods. In our numerical experiments, we use the state-of-the-art actor-critic algorithm twin-delayed deep deterministic policy gradient.

\subsubsection{Twin-Delayed Deep Deterministic Policy Gradient}\label{subsec:TD3}

In this work, we rely on the twin-delayed deep deterministic policy gradient (TD3) algorithm \cite{fujimoto2018addressing}. TD3 is an actor-critic algorithm learning a deterministic policy (the actor), and action-value function (the critic) by means of neural networks of parameters $\xi$ and $\theta$ respectively. TD3 is capable of handling continuous state and action spaces, making it a suitable candidate for controlling parametric PDEs. We indicate the parametrized policy by $\pi(s; \xi)$ and the action-value function by $Q(s, a;\theta)$.
The action-value function $Q(s, a;\theta)$ is updated via temporal-difference learning \cite{sutton2018reinforcement} as:
\begin{equation}
\begin{split}
        \mathcal{L}(\theta) &= \mathbb{E}_{s_t, a_t, s_{t+1}, r_t \sim \mathcal{M}}[(r_t + \gamma \bar{Q}(s_{t+1}, a_{t+1}; \bar{\theta}) - Q(s_t, a_t; \theta))^2] \\
        &= \mathbb{E}_{s_t, a_t, s_{t+1}, r_t \sim \mathcal{M}}[(r_t + \gamma \bar{Q}(s_{t+1}, \bar{\pi}(s_{t+1};\bar{\xi})+\epsilon; \bar{\theta}) - Q(s_t, a_t; \theta))^2],
\end{split}
\end{equation}
where the so-called target networks $\bar{Q}(s, a; \bar{\theta})$ and $\bar{\pi}(s;\bar{\xi})$ are copies of $Q(s, a;\theta)$ and $\pi(s; \xi)$, respectively, with parameters frozen, i.e. not updated in the backpropagation step, improving the stability of the training, $\epsilon \sim \text{clip}(\mathcal{N}(0, \bar{\sigma}), -c, c)$ is the noise added to estimate the action value in the interval $[-c, c]$ around the target action, and $\mathcal{M}$ is the memory buffer containing the samples collected by the agent while interacting with the environment.
To reduce the problem of overestimation of the target Q-values, TD3 estimates two value functions, namely $Q(s,a;\theta_1)$ and  $Q(s,a;\theta_2)$ and computes the target values as:
\begin{equation}
    r_t + \gamma \bar{Q}(s_{t+1}, a_{t+1}; \bar{\theta}) = r_t + \gamma  \min_{i=1, 2}\bar{Q}(s_{t+1},  a_{t+1}; \bar{\theta_i}).
\end{equation}

The target networks, parametrized respectively by $\bar{\theta}_1$, $\bar{\theta}_2$, and $\bar{\xi}$, are updated with a slower frequency than the actor and the critic according to:
\begin{equation}
\begin{split}
\bar{\theta}_1 &= \rho \theta_1 + (1-\rho)\bar{\theta}_1\\
\bar{\theta}_2 &= \rho \theta_2 + (1-\rho)\bar{\theta}_2\\
\bar{\xi} &= \rho \xi + (1-\rho)\bar{\xi}, \\
\end{split}
\label{target_nns_update}
\end{equation}
where $\rho$ is a constant determining the speed of the updates of the target parameters.

The action-value function $Q(s, a;\theta)$ is used to update the parameters of the deterministic policy $\pi(s;\xi)$. In particular, the gradient of the critic guides the improvements of the actor:
\begin{equation}
    \mathcal{L}(\xi) = \mathbb{E}_{s_t \sim \mathcal{M}}[-\nabla_{a} Q(s_t, \pi(s_t; \xi); \theta)].
\label{eq:policytd3}
\end{equation}

\subsection{Sparse Dictionary Learning}\label{subsec:SINDy}
Dictionary-learning methods are data-driven methods aiming to approximate a nonlinear function by finding a linear combination of candidate dictionary functions, e.g., polynomials of degree $d$ or trigonometric functions. Sparse dictionary-learning techniques additionally enforce sparsity and try to balance function complexity with approximation accuracy (identifying the smallest number of non-zero dictionary functions that can still accurately approximate the nonlinear function). In the context of learning and controlling dynamical systems from data, the sparse identification of nonlinear dynamics method (SINDy) \cite{brunton2016discovering} can discover governing equations by relying on sparse dictionary learning.

In particular, given a set of $N$ input data $S = [s_1, \cdots, s_N]$ and labeled output data $A = [a_1, \cdots, a_N]$, we want to identify the unknown function $f:\mathcal{S}\rightarrow\mathcal{A}$ such that $s = f(a)$. To find the best approximation of $f$, we construct a dictionary of candidate functions $\Theta(S)= [\theta_1(S), \cdots, \theta_D(S)]$. Given this dictionary, we can write the input-output relation as:
\begin{equation}
    A = \Theta(S)\Xi,
\end{equation}
where $\Xi$ is the matrix of coefficients to be fit. In contrast with neural networks, which pose no restriction on the function class that is approximated, the choice of dictionary restricts the possible functions that can be approximated and reduces the number of learnable parameters, in this case, the coefficients of the matrix $\Xi$. 

\subsection{Sparsifying Neural Network Layers with L$_0$ Regularization}\label{subsec:L0reg}

To sparsify the weight/coefficient matrix $\Xi$, the differentiable L$_0$ regularization method introduced in \cite{louizos2018learning} can be used. The method relaxes the discrete nature of L$_0$ to allow efficient and continuous optimization. 

Let $d$ be a continuous random variable distributed according to a distribution $p(d| \psi)$, where $\psi$ indicates the parameters of $p(d| \psi)$. Given a sample from $d \sim p(d|\psi)$, we can define:
\begin{equation}
    z= \min(1, \max(0, d)).
    \label{eq:hard-sigmoid-rectification}
\end{equation}
The hard-sigmoid rectification in Equation \eqref{eq:hard-sigmoid-rectification} allows the gate to be exactly zero. Additionally, we can still compute the probability of the gate being action, i.e., non-zero, by utilizing the cumulation distribuition function $P$:
\begin{equation}
    p(z\neq 0| \psi) = 1 - P(d \leq 0|\psi).
\end{equation}
We choose as candidate distribution a binary concrete \cite{maddison2016concrete, jang2016categorical}. 
We can now optimize the parameters $\psi$ of the distribution and introduce the L$_0$ regularization loss as:
\begin{equation}
    L_0(\psi) = \sum_{j=1}^{|\xi|} (1-P_{\bar{d}_j}(0|\psi)) = \sum_{j=1}^{|\xi|} \sigma(\log \alpha_j - \beta \log \frac{\gamma}{\zeta}),
\label{eq:L0_loss}
\end{equation}
where $\xi$ are the parameters of the model we want to sparsify and $\sigma$ corresponds to the sigmoid activation function. At test time, we can estimate the sparse parameters $\xi^0$ by:
\begin{equation}
    \begin{split}
        z &= \min(1, \max(0, \sigma(\log \alpha)(\zeta - \gamma) + \gamma)) , \, \\
        \xi^0 &= \xi \odot z, \,
    \end{split}
\end{equation}
where  $\odot$ corresponds to the element-wise product. 

In the context of sparse dictionary learning, replacing the L$_1$ norm with a L$_0$ norm has shown to be beneficial in the context of SINDy \cite{zheng2018unified, champion2020unified}. Moreover, recent work shows that replacing the truncated L$_1$ norm with the L$_0$ loss in Equation \eqref{eq:L0_loss} provides improved performance when combining dimensionality reduction using variational autoencoders \cite{kingma2013auto} with SINDy for discovering governing equations of stochastic dynamical systems~\cite{jacobs2023hypersindy}.

\section{Methodology}\label{sec:method}

\subsection{Problem Settings}\label{subsec:problem_settings}
In this work, we consider the problem of controlling continuous-time dynamical systems that evolve according to:
\begin{equation}
    \dot{y}(x, t) = f(y(x, t), u(x, t);\mu)
    \label{eq:dyn_system}
\end{equation}
where $x$ and $t$ represent the spatial and time domain variables, respectively, $y(x, t)=y \in \mathcal{Y}$ is the set of states, $\dot{y}(x, t)=\frac{dy(x, t)}{dt}$ is the time derivative of the state, $u(x, t)=u \in \mathcal{U}$ is the control input, and $\mu \in \mathcal{P}$ is a vector of the system parameters. We assume no knowledge of the operator-valued function $f: \mathcal{Y}\times\mathcal{A} \times \mathcal{P}\rightarrow \mathcal{Y}$, possibly containing high-order derivatives in the spatial domain variable $x$, but we can measure the evolution of the system at different discrete-time instants, resulting in a sequence of measurements $m_t, m_{t+1}, \cdots, m_{t+H}$. We assume that only a small finite number of sensors are available, therefore, measurements only contain a subset of the information of the full state of the system. Similarly, we assume a small finite number of actuators to control our systems. The limited number of sensors and actuators makes the PDE control problem extremely challenging.

The control problem is to steer the state of the PDE towards a desired target value with the minimum control effort. Therefore, we introduce the cost functional $J(\cdot, \cdot)$:
\begin{equation}
    J(y, u) = \frac{1}{2}
||y(x,t) - y_{\text{ref}}||^2 + \alpha ||u(x,t) - u_{\text{ref}}||^2, \,
\label{eq:costfunctional}
\end{equation}
where $y_{\text{ref}}$ and $u_{\text{ref}}$ indicate the reference values for the state and the control input, and $\alpha$ is a scalar positive coefficient balancing the contribution of the two terms. 

\subsection{Deep Reinforcement Learning with L$_0$-Sparse Polynomial Policies}
We introduce a sample efficient, interpretable, and robust DRL approach for efficient learning of optimal policies with a limited number of parameters for (approximately) solving the optimal control problem in Equation \ref{eq:costfunctional}. We devise a general method for policy approximation using dictionary learning with L$_0$ regularization to enforce sparsity. Our method can be used in different policy-gradient and actor-critic DRL algorithms without changing their policy-optimization procedure. 
Central to our method is a differentiable L$_0$ norm to enforce sparsity in the optimal policy, which makes the method compatible with gradient-based optimizations such as SGD or ADAM.

\subsubsection{TD3 Gradient with L$_0$-Sparse Polynomial Policy}\label{subsec:sparseTD3}
With reference to Figure \ref{fig:framework}, our sparse polynomial policy maps the state of the agent $s= [m, \mu]=[m_1, \cdots, m_n, \mu_1, \cdots, \mu_p]$ to control inputs $u$, where $m_i$ corresponds to the $i^{th}$-sensory measurement, and $\mu_i$ to the $i^{th}$-parameter value. The sparse polynomial policy is constructed as follows:
\begin{enumerate}
    \item We encode the measurements $m$ and parameters $\mu$ into a higher-dimensional space using a library of polynomial functions $\Theta$ of degree $d$, such that:  $[m, \mu] \xrightarrow{\Theta} [1, m, m^2, \dots, m^d, \dots, \mu,  \mu^2, \dots \mu^d]  =\Theta(s) = \tilde{s}$.
    \item We feed the polynomial features $\tilde{s}$ to a single-layer neural network $\pi(\tilde{s};\xi)$ to obtain the control action $a = \pi(\tilde{s};\xi) = \Theta(s)\Xi$, where the weights $\xi$ of the neural network corresponds to the learnable coefficient of the polynomial features and $\Xi$ is the matrix representation of $\xi$.
    \item We sparsify the weight matrix $\Xi$ using a binary mask $Z(\Psi)=\bm{z}(\psi)$, where $Z$ an  $\Psi$ indicate the matrix representation of the mask $\bm{z}$ and the parameters $\psi$, respectively, that are trainable using L$_0$ regularization (see Equation \eqref{eq:L0_loss} in Section \ref{subsec:L0reg}).
\end{enumerate}
The overall policy can be defined as:
\begin{equation}
\begin{split}
        a &= \pi(\tilde{s}; \xi, \psi) =  \Theta(s)\Xi  \odot Z(\Psi), \,\\
        a &= \tanh{(a)}, \, \\
        u &= g(a)\, ,
\end{split}
    \label{eq:polypolicy}
\end{equation}
where the hyperbolic tangent $\tanh{(a)}$ is used to clip the action values in $[-1, 1]$ and $g(\cdot)$ to a generic function modelling the actuation, e.g., using a sum of Gaussians as shown in Equation \eqref{eq:ks}. 

In our method, we simply replace the neural network-based policy of TD3 (see Section \ref{subsec:TD3}) with the sparse polynomial policy in Equation \eqref{eq:polypolicy}. Additionally, we redefine the TD3 training objectives (Equation \eqref{eq:policytd3}) to promote sparsity using L$_0$ regularization (Equation \eqref{eq:L0_loss}). The new training objective of the policy becomes:
\begin{equation}
    \mathcal{L}(\xi, \psi) = \mathbb{E}_{s_t \sim \mathcal{M}}[-\nabla_{a} Q(s_t, \pi(\tilde{s}; \xi, \psi); \theta)+ \lambda L_0(\psi)],
\label{eq:policypolytd3}
\end{equation}
where $\lambda$ is a scaling factor for the two loss terms. The complete algorithm is shown in Algorithm \ref{alg:sparseTD3} and the implementation is detailed in Appendix \ref{app:DRL_implementation}.
\begin{algorithm}
\caption{TD3 with L$_0$-Sparse Polynomial Policies}\label{alg:sparseTD3}
\begin{algorithmic}
\State Initialize  $Q(s,a;\theta_1)$, $Q(s,a;\theta_2)$, and $\pi(\tilde{s}; \xi, \psi)$ with random parameters $\theta_1, \theta_2, \xi, \psi$
\State Initialize target networks $\bar{\theta}_1 \leftarrow \theta_1$, $\bar{\theta}_2 \leftarrow \theta_2$, $\bar{\xi} \leftarrow \xi$, $\bar{\psi} \leftarrow \psi$
\State Initialize memory buffer $\mathcal{M}$
\State Select library of dictionary functions $\Theta$
\For{$e=1:E_{\max}$}
\State Initialize the system and get initial measurement $s_0 =[m_0,\mu]$
\For{$t=1:T_{\max}$}
\State Compute polynomial features $\tilde{s}_t = [1, m_t, m_t^2, \dots, m_t^d, \dots, \mu, \mu^2, \dots \mu^d] \xleftarrow{\Theta} [m_t, \mu]$
\State Sample action $a_t \sim \pi(\tilde{s}_t; \xi, \psi) + \epsilon$, where $ \epsilon \sim \mathcal{N}(0, \sigma)$
\State Observe reward $r$ and new measurement $[m_{t+1},\mu]$
\State Store tuple $(s_t, a, r, s_{t+1})$ in $\mathcal{M}$
\State
\If{train models}
\State Sample mini-batch $(s_t, a, r, s_{t+1})$ from $\mathcal{M}$
\State Compute polynomial features $\tilde{s}_{t+1}, \tilde{s}_{t} \xleftarrow{\Theta} [m_{t+1}, \mu], [m_{t}, \mu]$
\State $a_{t+1} \leftarrow \bar{\pi}(\tilde{s}_t; \bar{\xi}, \bar{\psi}) +  \epsilon$, where $\epsilon \sim \text{clip}(\mathcal{N}(0, \bar{\sigma}), -c, c)$
\State $q_t \leftarrow r_t + \gamma  \min_{i=1, 2}\bar{Q}(s_{t+1},  a_{t+1}; \bar{\theta_i})$
\State Update critic parameters according:
\State $\mathcal{L}(\theta) = \mathbb{E}_{s_t, a_t, s_{t+1}, r_t \sim \mathcal{M}}[(q_t - Q(s_t, a_t; \theta))^2]$
\If{train actor}
\State Update policy parameters according to:
\State $    \mathcal{L}(\xi, \psi) = \mathbb{E}_{s_t}[-\nabla_{a} Q(s_t, \pi(\tilde{s}_t; \xi, \psi); \theta)+ \lambda L_0(\psi)]$
\State Update target networks
\State $\bar{\theta}_1 = \rho \theta_1 + (1-\rho)\bar{\theta}_1$
\State $\bar{\theta}_2 = \rho \theta_2 + (1-\rho)\bar{\theta}_2$
\State $\bar{\xi} = \rho \xi + (1-\rho)\bar{\xi}$
\State $\bar{\psi} = \rho \psi + (1-\rho)\bar{\psi}$
\EndIf
\EndIf
\State
\EndFor
\EndFor
\end{algorithmic}
\end{algorithm} 

Our method allows us to retain the expressive power of the neural networks for approximating the value function, i.e., the expected discounted return of the state-action pairs which may be very difficult to approximate, while exploiting the simpler structure and interpretability of polynomials for representing a policy. The algorithm is capable of learning complex policies with limited number of parameters, but, differently from a neural network-based policy, it returns interpretable polynomial expressions that can be used, for stability or robustness analysis. 
Eventually, it is worth mentioning that our approach can be easily adapted to any policy-gradient or actor-critic DRL algorithms such as DDPG, PPO, and SAC.

\section{Results}\label{sec:results}
To test the validity of our approach, we study two control problems of different parametric PDEs, namely the Kuramoto-Sivashinsky (KS) and the Convection-Diffusion-Reaction (CDR) PDE.
We compare our sparse L$_0$-sparse polynomial TD3 agent, with polynomial degree $d=3$, with:
\begin{itemize}
    \item the TD3 agent with value function and policy represented by neural networks. We utilize the default TD3 hyperparameters and architecture from \cite{fujimoto2018addressing}. Details of the implementation can be found in Appendix \ref{app:DRL_implementation},
    \item the L$_1$-sparse polynomial TD3 agent, where we replace the L$_0$ regularization with an L$_1$ regularization (see details in Appendix \ref{app:polyl1}), and
    \item the TD3 agent without the parameter $\mu$ in the agent state  $s$.
\end{itemize}

\subsection{Kuramoto-Sivashinsky PDE}
\begin{figure}[h!]
    \centering
    \includegraphics[width=1.0\textwidth]{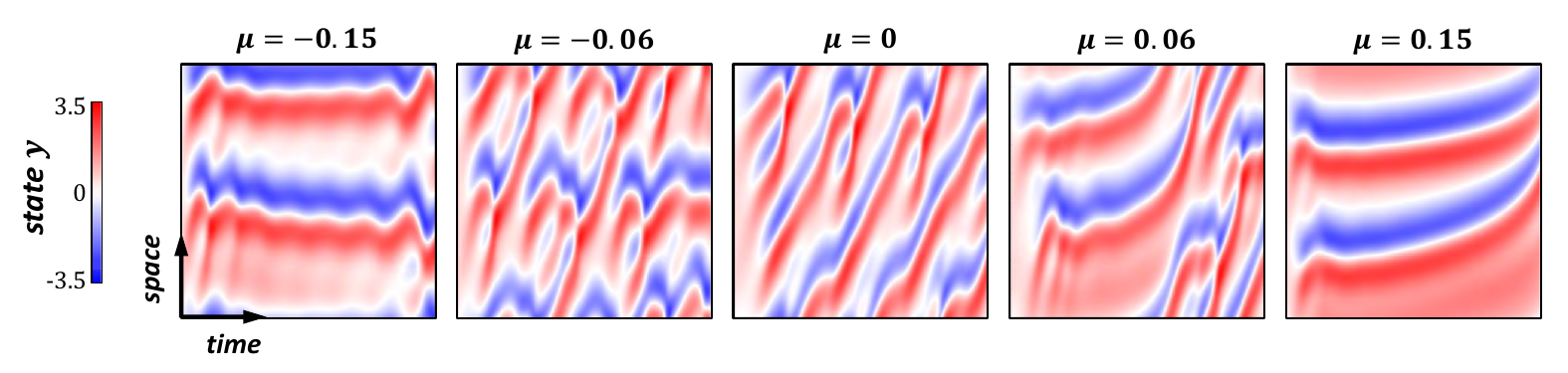}
        \caption{Example of solutions of the Kuramoto-Sivashinsky PDE for different values of the parameter $\mu$.}
        \label{fig:example_solution_ks}
\end{figure}

The KS is a nonlinear 1D PDE describing pattern and instability in fluid dynamics, plasma physics, and combustion, e.g. the diffusive-thermal instabilities in a laminar flame front \cite{kudryashov1990exact}. Similarly to \cite{peitz2023distributed}, we write the KS PDE with state $y(x, t)=y$ with the addition of a parametric spatial cosine term, breaking the spatial symmetries of the equation and making the search for the optimal control policy more challenging:
\begin{equation}
\begin{split}
     \frac{\partial y}{\partial t} + y\frac{\partial y}{\partial x} + \frac{\partial^2 y}{\partial x^2} + \frac{\partial^4 y}{\partial x^4} + \mu \cos{(\frac{4\pi x}{L})} &=  u(x, t) \\
     u(x, t) &=\sum_{i=1}^{N_a}a_i \psi(x, c_i)  \\
     \psi(x, c_i) &=\frac{1}{2}\exp(-{(\frac{x-c_i}{\sigma})}^2) \\
\end{split}
\label{eq:ks}
\end{equation}
where $u(x, t)$ is the control input function with $a_i \in [-1, 1]$, $\psi(x, c_i)$ is a Gaussian kernel of mean $c_i$ and standard deviation $\sigma=0.8$, $\mu \in [-0.25, 0.25]$ is the parameter of interest of the system, and $\mathcal{D}=[0, 22]$ is the spatial domain with periodic boundary conditions. To numerically solve the PDE, we discretize the domain with $N_x=64$, analogously to \cite{peitz2023distributed}. We assume to have $8$ equally-spaced actuators and sensors. The state of the agent is composed of $8$ sensory readings and the scalar value of the parameter, making each agent state $s=[m_1, \cdots, m_8, \mu]$ of dimension $N_s=9$. Examples of solutions for different values of $\mu$ are shown in Figure \ref{fig:example_solution_ks}.

We aim to steer the state of the PDE $y$ towards a desired reference state with the minimum control effort. Therefore, analogously to Equation \eqref{eq:costfunctional}, we can write the reward function as:
\begin{equation}
\begin{split}
        R(y, u) &= -J(y, u) = -(c_1 + \alpha c_2) \\ &= -\Big(\frac{1}{N_x}\sum_{k=1}^{N_x}(y_{k,t} - y_{\text{ref}})^2 + \alpha \frac{1}{N_a} \sum_{j=1}^{N_a}  (u_{j,t} - u_{\text{ref}})^2\Big)\\
\end{split}
\label{eq:rew}
\end{equation}
where $y_{k,t}, u_{j,t}$ represent respectively the $n^{\text{th}}, j^{\text{th}}$ component of the discretized state vector $y_t$ and control input $u_t$ at the timestep $t$, $y_{\text{ref}}=0.0$, $u_{\text{ref}} = 0.0$, and $\alpha=0.1$. The choice of $\alpha=0.1$ is dictated by the need for balancing the contribution of the state-tracking cost $c_1$ and the control-effort cost $c_2$. In particular, in our experiments we prioritize steering the system state to the reference over the minimization of the injected energy.

We train the control policies by randomly sampling a value of the parameter $\mu$ at the beginning of each training episode. We choose $\mu \in [-0.2, -0.15,  -0.1, -0.05, 0.0, 0.05, 0.1, 0.15, 0.2]$. To test the generalization abilities of the policies, we evaluate their performance on unseen and randomly-sampled parameters from $[-0.25, 0.25]$. Additionally, to test the robustness of the policies, during the evaluation we add noise to the sensory readings $m_t = m_t + \epsilon$ with $\epsilon\sim \mathcal{N}(0, \sigma I)$ and $\sigma=0.25$.

\begin{figure}[h!]
    \centering
    \includegraphics[width=1.0\textwidth]{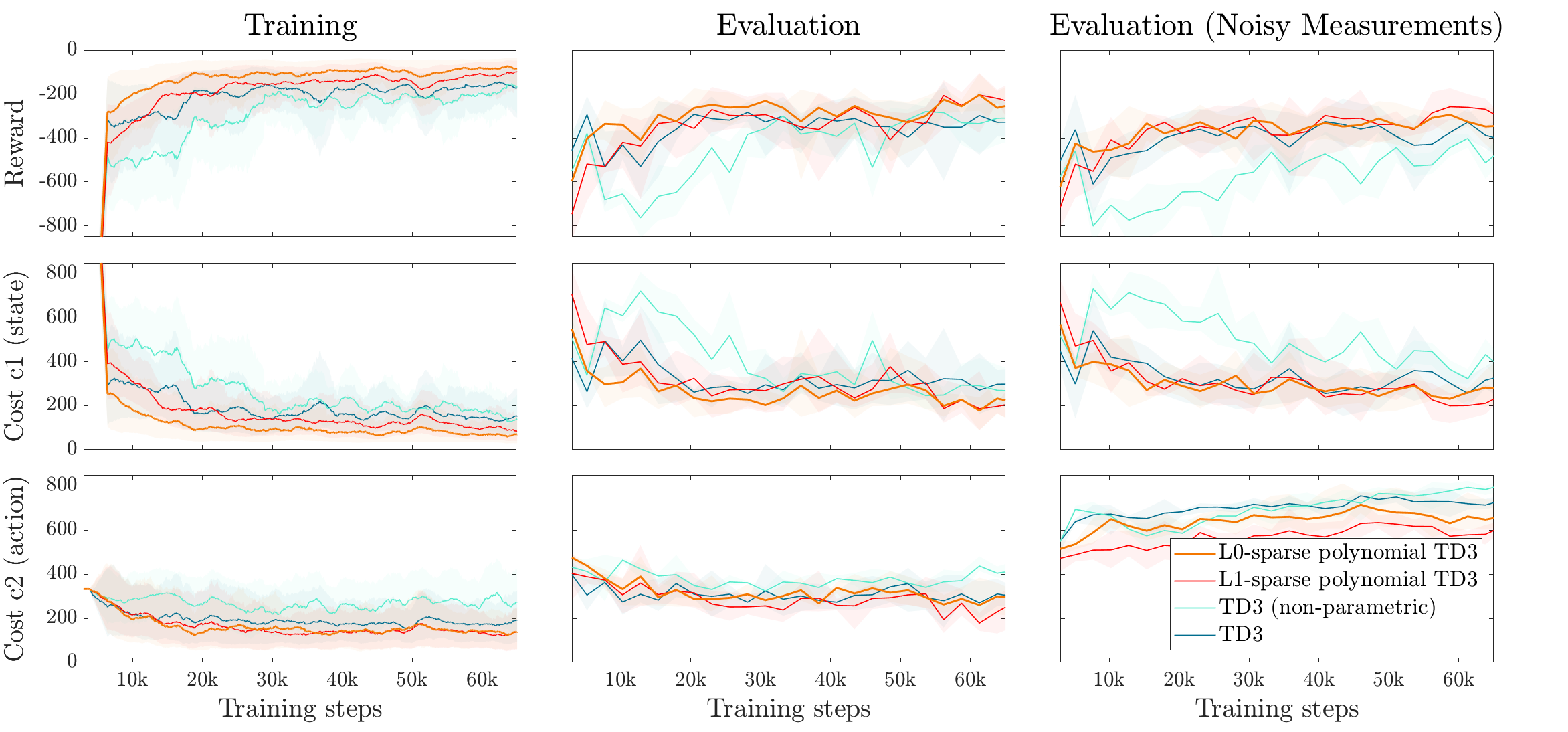}
        \caption{Training and evaluation (with and without noise on the measurements) results of the Kuramoto-Sivashinsky PDE control problem. The plots show mean (solid line) and standard deviation (shaded area) of four different random seeds ($1, 7, 92, 256$).}
        \label{fig:results_ks}
\end{figure}

In Figure \ref{fig:results_ks}, we show training and evaluation results of the different methods. In particular, we show the reward (see Equation \eqref{eq:rew}), the state cost $c_1$, and the action cost $c_2$ over training and evaluation with and without noise on the measurements. Our L$_0$-sparse polynomial TD3 agent achieves the highest reward and the lowest total cost over training and in evaluation to unseen parameters. Therefore, enforcing sparsity, either through L$_0$ or L$_1$ regularization, improves the performance of the agents even in the presence of a large amount of noise on the measurements. 

\begin{figure}[h!]
     \centering
    \includegraphics[width=1.0\textwidth]{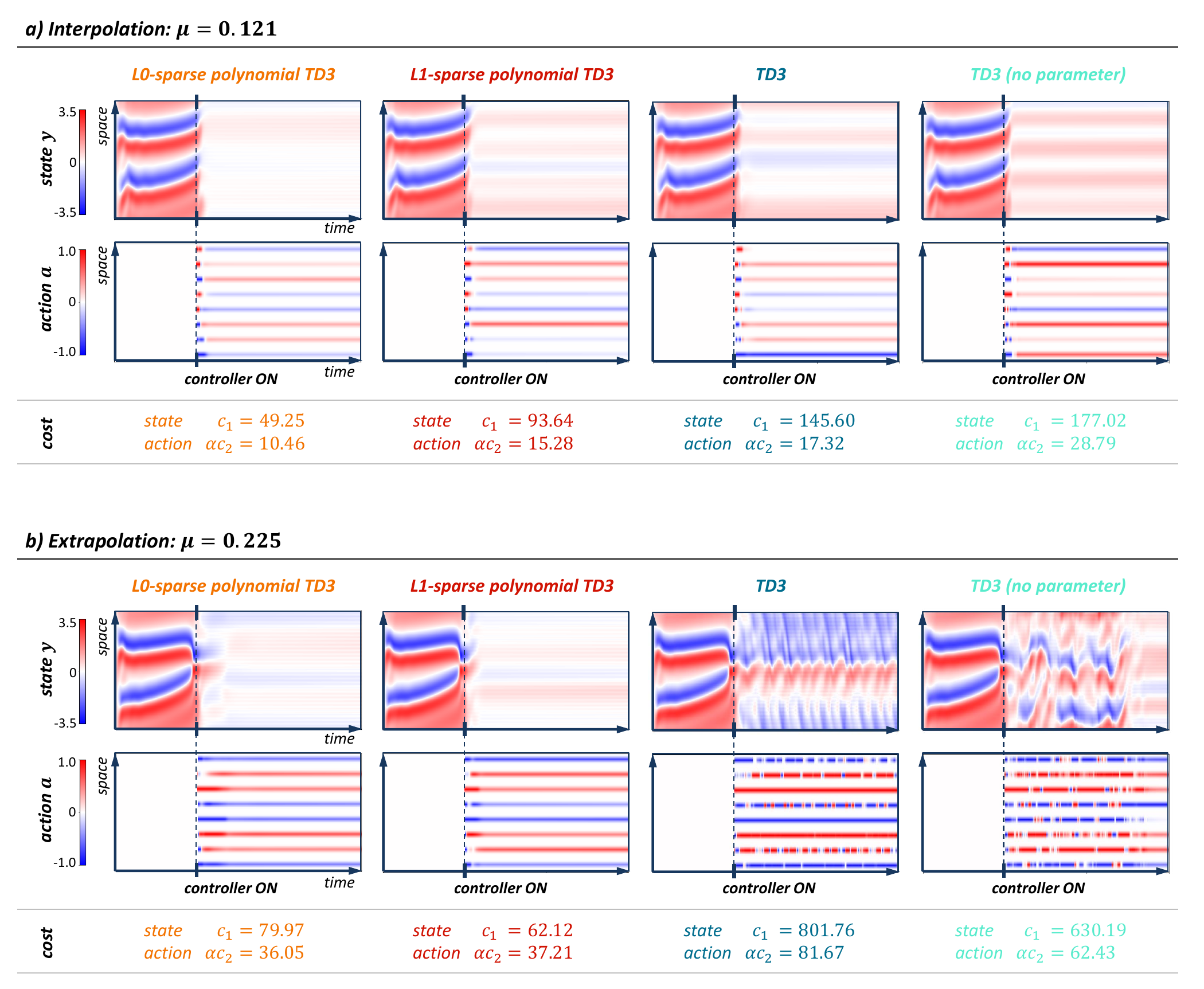}
        \caption{Optimal control policy performance on unseen instances of the parameter $\mu$, i.e., a) $\mu=0.121$ (interpolation), and b) $\mu=0.225$ (extrapolation). Additionally, for each method we report the state cost $c_1$ and the scaled action cost $\alpha c_2$.}
        \label{fig:results_ks_id}
\end{figure}

In Figure \ref{fig:results_ks_id}, we show examples of optimal control solutions for the KS PDE for two different values of the parameter $\mu$ using the different DRL control policies. We highlight that those specific instances of the parameter $\mu$ were never seen by the agents during training. In particular, in Figure \ref{fig:results_ks_id}a), we select a value of $\mu$ within the training range to test the \textit{interpolation} capabilities of the controllers, while in Figure \ref{fig:results_ks_id}b), we select a value of $\mu$ outside the training range to test the \textit{extrapolation} capabilites of the controllers. 
In interpolation regimes, the L$_0$-sparse polynomial TD3 outperforms the L$_1$-sparse polynomial TD3, the DNN-based TD3, and the DNN-based TD3 without the parameter $\mu$ in the agent state $s$. In extrapolation regimes, we observe similar performance of the L$_0$-sparse and the L$_1$-sparse polynomial TD3, drastically outperforming their DNN-based counterparts.
First, these results show the importance of observing the parameter $\mu$ of the PDE, and second, they show the importance of enforcing sparsity when developing controllers for parametric PDEs capable of generalizing to new instances of the parameter.

Compared to black-box neural networks, our method learns polynomial controllers that can be written in closed form, and that can be analyzed and interpreted. In the case of the KS PDE, starting from an observation vector of dimension 9 (8-dimensional sensory measurements and 1-dimensional parameter) and a polynomial degree $d=3$, we obtain a polynomial with 220 coefficients (we allow cross products between the different variables of the vector). Since we have 8 possible control inputs, the total number of learnable parameters is 220*8=1760, excluding the L$_0$ mask coefficients. With only 1760 learnable coefficients, our method drastically reduces the number of learnable parameters compared to the standard 3-layers neural network used by TD3 (and in general by DRL algorithms). Examples of optimal sparse control policies are:
\begin{equation}
\begin{cases} 
a_4 = \tanh{ (-7.524m_4 +0.226m_3^2 -2.128 m_3 m_7 -2.659 m_8\mu -0.691m_1 \mu^2 -0.317 \mu^3)} \\ 
a_8 = \tanh{(0.234 +1.101m_5 - 1.853m_7 -2.659m_8 -6.481m_8^3 -2.993m_8\mu^2 -0.567\mu^3)}  \\ 
\end{cases} 
\label{eq:polyTD3_ks}
\end{equation}
For Equation \eqref{eq:polyTD3_ks}, it is possible to notice that the control laws are not only interpretable and sparse, i.e., composed of a few terms only, but they also strongly depend on local measurements, $m_4, m_8$ respectively, and on the parameter $\mu$. Additionally, the control laws present nonlocal-interaction terms, e.g., in the first equation the action $a_4$ depends on the measurement $m_1$, indicating nonlocal information is essential to derive optimal control strategies and that the proposed policy-optimization approach with L$_0$ regularization is capable of unveiling these relations. 

\subsection{Convection-Diffusion-Reaction PDE}
\begin{figure}[h!]
    \centering
    \includegraphics[width=1.0\textwidth]{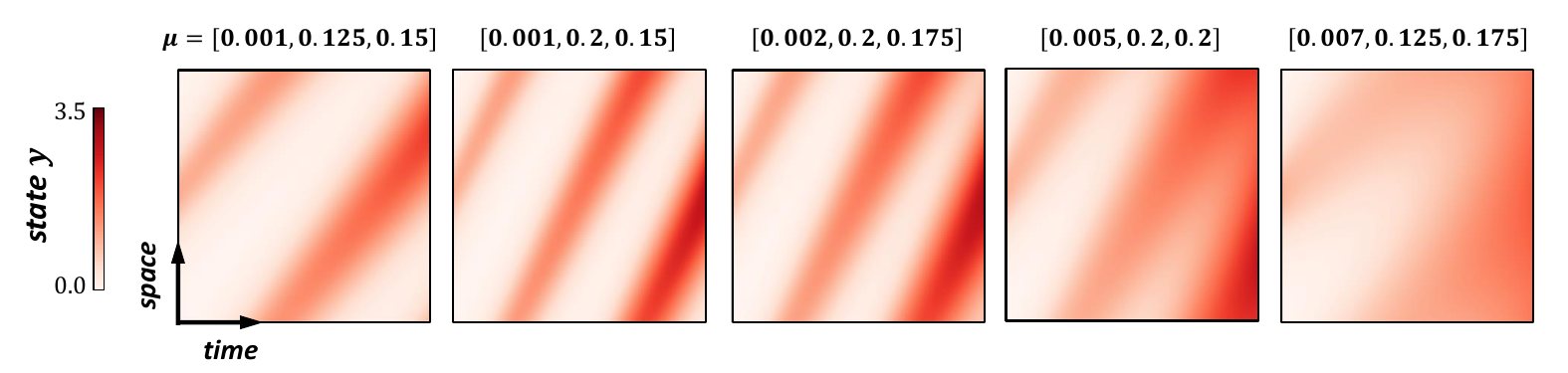}
        \caption{Example of solutions of the Convection-Reaction-Diffusion PDE for different values of the parameter vector $\mu=[\nu, c, r]$.}
        \label{fig:example_solution_cdr}
\end{figure}

The CDR PDE describes the transfer of energy, or physical quantities withing a system due to convection, diffusion, and reaction processes. The CDR PDE with state $y(x, t)=y$ can be written as:
\begin{equation}
\begin{split}
     \frac{\partial y}{\partial t} + c\frac{\partial y}{\partial x} - \nu \frac{\partial^2 y}{\partial x^2} -ry &=  u(x, t) \\
     u(x, t) &=\sum_{i=1}^{N_a}a_i \psi(x_i, c_i),  \\
\end{split}
\label{eq:cdr}
\end{equation}
where $\psi(x_i, c_i)$ is a Gaussian kernel with $a_i \in [-0.25, 0.25]$, similar to the control input function in the KS PDE problem, $c \in [0.1, 0.35]$ is the convection velocity, $\nu \in [0.001, 0.008]$ is the diffusivity constant, $r \in [0.1, 0.35]$ is the reaction constant, and $\mathcal{D}=[0, 1]$ is the spatial domain with periodic boundary conditions. To solve the PDE, we assume a spatial discretization $N_x=200$. We use $8$ sensors and $8$ actuators. In this PDE control problem, we have three parameters $\mu=[\nu, c, r]$, therefore, the agent state $s=[m_1, \cdots, m_8, \bm{\mu}]$ is of dimension $N_s=11$. The implementation details for simulating the CDR PDE \cite{zhang2023controlgym} are provided in Appendix \ref{app:cdr}, and examples of solutions for different values of $\mu$ are shown in Figure \ref{fig:example_solution_cdr}.

Similarly to KS, the control problem is to steer the state towards a desired target value with the minimum control effort. Therefore, analogously to Equation \eqref{eq:rew}, we can write the reward function as:
\begin{equation}
\begin{split}
        R(y, u) &= -J(y, u) = -(c_1 + \alpha c_2) \\ &= -\Big(\frac{1}{N_x}\sum_{k=1}^{N_x}(y_{k,t} - y_{\text{ref}})^2 + \alpha \frac{1}{N_a} \sum_{j=1}^{N_a}  (u_{j,t} - u_{\text{ref}})^2\Big)\\
\end{split}
\end{equation}
where $y_{k,t}, u_{j,t}$ represent respectively the $k^{\text{th}}, j^{\text{th}}$ component of the full state vector $y_t$ and control input $u_t$, $y_{\text{ref}}=0.0$, $u_{\text{ref}} = 0.0$, and $\alpha=0.1$.

We train the control policies by randomly sampling values of the parameters $c$ and $r$ from $[0.1, 0.125, 0.15, 0.175, 0.2]$ and $\nu$ from $[0.001, 0.002,  0.003,  0.004,  0.005,  0.006,  0.007]$ at the beginning of each training episode. To test the generalization ability of the policies, we evaluate their performance on unseen and randomly sampled parameters within (interpolation) and without (extrapolation) the training range. Additionally, to test the robustness of the policies, we add noise to the measurements $m_t = m_t + \epsilon$ with $\epsilon\sim \mathcal{N}(0, \sigma I)$ and $\sigma=0.25$ during the evaluation.

\begin{figure}[h!]
     \centering
    \includegraphics[width=1.0\textwidth]{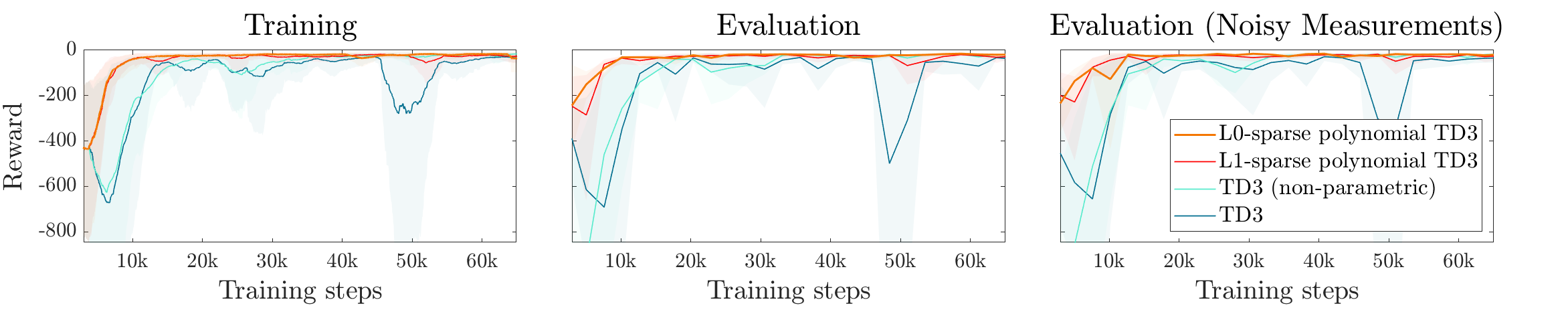}
        \caption{Training and evaluation results (with and without noise on the measurements) on the problem of control of the Convection-Diffusion-Reaction PDE. The plots show mean and standard deviation of four different random seeds ($1, 7, 92, 256$).}
        \label{fig:results_cdr}
\end{figure}

In Figure \ref{fig:results_cdr}, we show training and evaluation results of the different methods. Our L$_0$-sparse polynomial TD3 achieves higher reward over training and in evaluation to unseen parameters. It is worth noticing that enforcing sparsity improves the performance of the control policies even in the presence of large noise on the measurements. 

\begin{figure}[h!]
     \centering
     \includegraphics[width=1.0\textwidth]{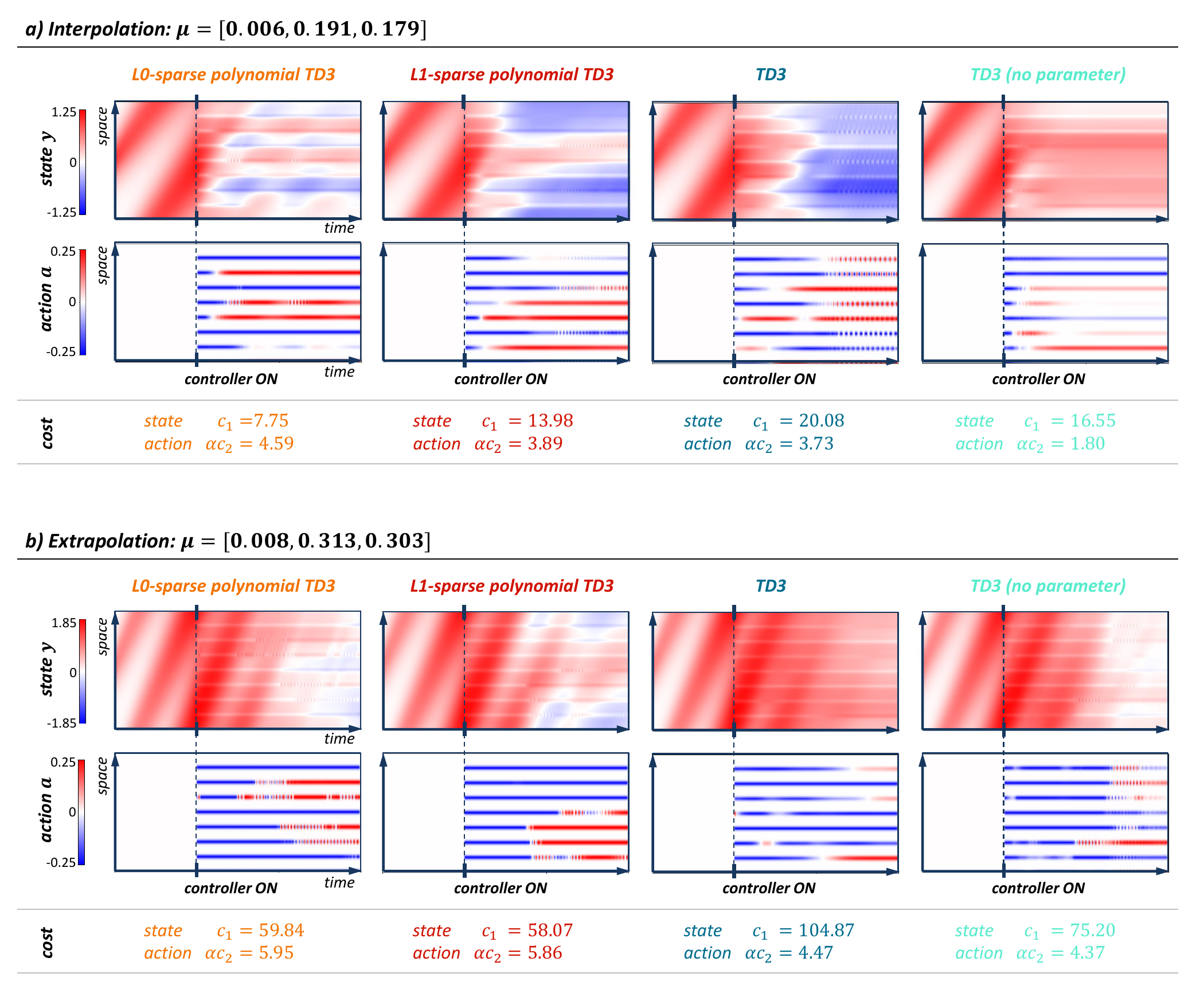}
        \caption{Optimal control policy performance on unseen instances of the parameter $\mu$, i.e., a) $\mu=[0.006, 0.191, 0.179$ (interpolation), and b) $\mu=[0.008, 0.313, 0.303]$ (extrapolation). Additionally, for each method we report the state cost $c_1$ and the scaled action cost $\alpha c_2$.}
        \label{fig:solutions_cdr}
\end{figure}

In Figure \ref{fig:solutions_cdr}, we show examples of optimal control solutions for the CDR PDE for different values of the parameter vector $\mu$ using the different control policies. We highlight that those specific instances of the parameter vector $\mu$ were never seen by the agents during training. In particular, in Figure \ref{fig:solutions_cdr}a) we select a value of $\mu$ within the training range to test the \textit{interpolation} capabilities of the controllers, while in Figure \ref{fig:solutions_cdr}b), we select a value of $\mu$ outside the training range to test the \textit{extrapolation} capabilites of the controllers. 
Even for the CDR PDE, in interpolation regimes, the L$_0$-sparse polynomial TD3 outperforms the L$_1$-sparse polynomial TD3, the DNN-based TD3, and the DNN-based TD3 without the parameter $\mu$ in the agent state $s$. In extrapolation regimes, we observe similar performance of the L$_0$-sparse and the L$_1$-sparse polynomial TD3, drastically outperforming their DNN-based counterparts.
Due to the limited actuation and sensing, the CDR PDE was found to be more challenging to control than the KS PDE, making the study of its observability and controllability in parametric settings an interesting direction for future research.

\section{Discussion and Conclusion}\label{sec:discussion}

Despite the plethora of DRL and optimal control approaches for solving PDE control problems, these methods often focus on solving the control problems for fixed instances of the PDE parameters. However, learning optimal control policies that can generalize to unseen instances of the parameters without further retraining/optimizing is crucial for reducing the computational cost of solving these control problems and for tackling complex real-world applications. 

In this work, we introduce a DRL framework for addressing the problem of efficient control of parametric PDEs by learning sparse polynomial control policies. To reduce the number of coefficients of the polynomials, and consequently learn sparse polynomial control policies, we use dictionary learning and differentiable L$_0$ regularization. We test our method on the challenging problem of controlling a parametric Kuramoto-Sivashinsky PDE (Equation \eqref{eq:ks}), and a parametric convection-diffusion-reaction PDE (Equation \eqref{eq:cdr}). While the control of this PDE has been the core focus of several works, to the best of our knowledge, the problem has not been addressed in parametric settings with large variations of the parameters. 
Also, recent work has not focused on learning optimal control policies with limited data (data collected at a limited number of parameters) and policies that can generalize to unseen parameters.
Compared to recent work \cite{peitz2023distributed} introducing a multi-agent DRL framework for controlling the Kuramoto-Sivashinsky PDE with parameters $\mu \in [0.0, 0.02]$, we extend the range of parameters by more than one order of magnitude. Additionally, our framework learns control policies that require significantly less control effort, making them more suitable for energy-constrained real-world applications.

An important advantage of learning DRL policies with sparse polynomials is that interpretable control strategies can be obtained, compared to black-box DNNs. 
Polynomial control policies composed of only a few active terms allow for robustness and stability analysis, which are critical when deploying the policies in real-world and safety-critical applications. 
Also, sparse policies are more robust to noise, making them suitable for solving real-world problems where measurements are often corrupted by noise. 

In this work, we use a dictionary of polynomial functions of degree $d=3$. However, our approach is not limited to polynomials. Prior knowledge of the optimal control policy can be used to choose the library accordingly. Additionally, we can incorporate constraints by removing or adding specific terms to the library, which is difficult when using DNN policies. 

Our work is inspired by the recent SINDy-RL algorithm that introduces the idea of dictionary learning with DRL \cite{zolman2024sindy}. SINDy-RL is a unifying framework that combines SINDy and DRL to learn representations of the dynamics model, reward function, and control policy. The three main differences between SINDy-RL and our work are:
\begin{enumerate}
    \item SINDy-RL learns a DNN control policy and then approximates the optimal policy using SINDy (i.e. sparse dictionary learning) to obtain a sparse polynomial control policy. Approximating the DNN policy a posteriori may introduce an additional approximation error that can deteriorate the policy performance. Our control policy is sparse and polynomial by construction, without utilizing a DNN.
    \item SINDy uses sequentially thresholded least squares to find sparse solutions, while our method uses a differentiable L$_0$-norm. 
    \item SINDy-RL is a model-based DRL method, learning the environment dynamics in addition to the policy and reward function. Our work focuses on a model-free approach that can generalize across different parametric PDEs.
\end{enumerate}

Many control problems require learning a sequence of decisions from high-dimensional inputs. A drawback of using a library of polynomials to pre-process the observations of the DRL agent is the limited scaling of the number of features with respect to the dimension of the input and the degree of the polynomial. However, this problem can be alleviated by finding compact and low-dimensional representations of the measurements. Extending our method to use machine learning-based dimensionality-reduction techniques, such as autoencoders, is an exciting direction for future research.

Our method can generalize to unseen parameters of the PDE, even when trained on a small number of instances of the parameters. This opens up exciting avenues of future work, exploring the generalization capability and data-efficiency of our DRL method in complex real-world dynamical systems.

%Bibliography
\bibliographystyle{unsrt}  
\bibliography{references}  

\appendix
\section{PDE Implementation}
\subsection{Kuramoto-Sivashinsky PDE}\label{app:ks}
We translate into Python code, the Julia implementation of the KS PDE proposed in \cite{peitz2023distributed}. The parameters of our simulations are reported in Table \ref{Tab:app_ks}.
\begin{table}[h!]
\begin{center}
\caption{Parameters of the KS PDE simulations.}
\begin{tabular}{||c c||} 
 \hline
 Parameter & Value \\ [0.5ex] 
 \hline\hline
 Domain size ($L$) & 22 \\ 
 \hline
 Spatial discretization ($N_x$) & 64 \\ 
  \hline
 T & 300s  \\
 \hline
 Control starts & 100s \\
 \hline
 $\Delta t$ & 0.1s \\
 \hline
  $\Delta t$ controller & 0.2s \\
  \hline
  $\alpha$ & 0.1 \\
 \hline
 Sensors & 8 \\
  \hline
 Parameter & 1 \\
 \hline
 State dimension ($N_s$) & 9 \\
 \hline
 Actuators ($N_a$) & 8 \\
 \hline
 Gaussian actuators $\sigma$ & 0.8 \\ [0,5ex] 
 \hline
\end{tabular}
\label{Tab:app_ks}
\end{center}
\end{table}

\subsection{Convection-Diffusion-Reaction PDE}\label{app:cdr}
For the CDR PDE, we improve the implementation of \cite{zhang2023controlgym} by:
\begin{itemize}
    \item replacing the support function of the actuators from a square wave to a Gaussian, and
    \item starting the controller at timestep 50, instead of 0, to allow the dynamics to evolve sufficiently, making the control task more challenging.
\end{itemize}
The parameters used in our simulations are reported in Table \ref{Tab:app_cdr}.
\begin{table}[h!]
\begin{center}
\caption{Parameters of the CDR PDE simulations.}
\begin{tabular}{||c c||} 
 \hline
 Parameter & Value \\ [0.5ex] 
 \hline\hline
 Domain size ($L$) & 1 \\ 
 \hline
 Spatial discretization ($N_x$) & 200 \\ 
  \hline
 T & 15s  \\
  \hline
 Control starts & 5s \\
 \hline
 $\Delta t$ & 0.1s \\
 \hline
  $\Delta t$ controller & 0.2s \\
  \hline
  $\alpha$ & 0.1 \\
 \hline
 Sensors  & 8 \\
 \hline
 Parameters & 3 \\
 \hline
 State dimension ($N_s$) & 11 \\
 \hline
 Actuators ($N_a$) & 8 \\
 \hline
 Gaussian actuators $\sigma$ & 2.5 \\ [0.5ex] 
 \hline
\end{tabular}
\label{Tab:app_cdr}
\end{center}
\end{table}

\section{Deep Reinforcement Learning Algorithms}\label{app:DRL_implementation}

\subsection{$\text{L}_1$-Sparse Polynomial TD3}\label{app:polyl1}
Analogously to the L$_0$-sparse polynomial TD3, the L$_1$-sparse polynomial TD3 replaces the neural network-based policy of TD3 (see Section \ref{subsec:TD3}) with a polynomial policy. However, we promote sparsity using L$_1$ regularization. The new training objective of the policy becomes:
\begin{equation}
    \mathcal{L}(\xi) = \mathbb{E}_{s_t, \mu \sim \mathcal{M}}[-\nabla_{a} Q(s_t, \mu, \pi(\tilde{s}; \xi); \theta)+ \lambda L_1(\xi)],
\label{eq:L1policypolytd3}
\end{equation}
where $\lambda$ is a scaling factor for the two loss terms and $L_1(\xi)=||\xi||_1$. The complete algorithms is shown in Algorithm \ref{alg:sparseTD3L1}.
\begin{algorithm}
\caption{TD3 with L$_1$-Sparse Polynomial Policies}\label{alg:sparseTD3L1}
\begin{algorithmic}
\State Initialize  $Q(s,a;\theta_1)$, $Q(s,a;\theta_2)$, and $\pi(\tilde{s}; \xi)$ with random parameters $\theta_1, \theta_2, \xi$
\State Initialize target networks $\bar{\theta}_1 \leftarrow \theta_1$, $\bar{\theta}_2 \leftarrow \theta_2$, $\bar{\xi} \leftarrow \xi$
\State Initialize memory buffer $\mathcal{M}$
\State Select library of dictionary functions $\Theta$
\For{$e=1:E_{\max}$}
\State Initialize the system and get initial measurement $s_0 =[m_0,\mu]$
\For{$t=1:T_{\max}$}
\State Compute polynomial features $\tilde{s}_t = [1, m_t, m_t^2, \dots, m_t^d, \dots, \mu, \mu^2, \dots \mu^d] \xleftarrow{\Theta} [m_t, \mu]$
\State Sample action $a_t \sim \pi(\tilde{s}_t; \xi) + \epsilon$, where $ \epsilon \sim \mathcal{N}(0, \sigma)$
\State Observe reward $r$ and new measurement $[m_{t+1},\mu]$
\State Store tuple $(s_t, a, r, s_{t+1})$ in $\mathcal{M}$
\State
\If{train models}
\State Sample mini-batch $(s_t, a, r, s_{t+1})$ from $\mathcal{M}$
\State Compute polynomial features $\tilde{s}_{t+1}, \tilde{s}_{t} \xleftarrow{\Theta} [m_{t+1}, \mu], [m_{t}, \mu]$
\State $a_{t+1} \leftarrow = \bar{\pi}(\tilde{s}_t; \bar{\xi}) +  \epsilon$, where $\epsilon \sim \text{clip}(\mathcal{N}(0, \bar{\sigma}), -c, c)$
\State $q_t \leftarrow r_t + \gamma  \min_{i=1, 2}\bar{Q}(s_{t+1},  a_{t+1}; \bar{\theta_i})$
\State Update critic parameters according:
\State $\mathcal{L}(\theta) = \mathbb{E}_{s_t, a_t, s_{t+1}, r_t \sim \mathcal{M}}[(q_t - Q(s_t, a_t; \theta))^2]$
\If{train actor}
\State Update policy parameters according to:
\State $    \mathcal{L}(\xi, \boldsymbol{\psi}) = \mathbb{E}_{s_t}[-\nabla_{a} Q(s_t, \pi(\tilde{s}_t; \xi); \theta)+ \lambda L_1(\xi)]$
\State Update target networks
\State $\bar{\theta}_1 = \rho \theta_1 + (1-\rho)\bar{\theta}_1$
\State $\bar{\theta}_2 = \rho \theta_2 + (1-\rho)\bar{\theta}_2$
\State $\bar{\xi} = \rho \xi + (1-\rho)\bar{\xi}$
\EndIf
\EndIf
\State
\EndFor
\EndFor
\end{algorithmic}
\end{algorithm}

\subsection{Hyperparameters}\label{app:DRL_hyper}

In Table \ref{Tab:td3_hyper}, we report the (hyper)parameters used in the DRL algorithms, i.e., TD3, L0-sparse polynomial TD3, and L$_1$-sparse polynomial TD3, respectively.
\begin{table}[h!]
\begin{center}
\caption{Parameters of the algorithm.}
\begin{tabular}{||c c c c||} 
 \hline
 Parameter & TD3 & L$_0$ poly. TD3 & L$_1$ poly. TD3 \\ [0.5ex] 
 \hline\hline
 batch size &  256 & 256 & 256 \\ 
 \hline
  hidden layer size & 256  & 256  & 256  \\
 \hline
 learning rate & $3e-4$  & $3e-4$ & $3e-4$ \\
  \hline
 $\tau$ & 0.005 & 0.005 & 0.005\\
 \hline
 discount factor $\gamma$ & 0.99 & 0.99 & 0.99 \\ 
 \hline
 regularization coefficient $\lambda$ &  -  &  0.0005  &  0.005\\ [0.5ex] 
  \hline
\end{tabular}
\label{Tab:td3_hyper}
\end{center}
\end{table}

\section{Control of PDE as a Block Markov Decision Process}\label{app:MDP_formuation}

In particular, a BMDP is a tuple $\langle \mathcal{S}, \mathcal{A}, \mathcal{O}, T, \Omega, \mathcal{R} \rangle$ where $\mathcal{S}$ is the set of unobservable states, $\mathcal{A}$ is the set of actions, $\mathcal{O}$ is the observation space (here assumed Markovian), $T: \mathcal{S} \times \mathcal{S} \times \mathcal{A} \longrightarrow [0,1]; \ (s',s,a) \longmapsto T(s',s,a)=p(s'|s,a)$ is the transition function, $R: \mathcal{S} \times \mathcal{A} \longrightarrow \mathbb{R}; \ (s,a)\longmapsto R(s,a)$ is the reward function, and $\Omega: \mathcal{O} \times \mathcal{S}\times \mathcal{A} \longrightarrow [0,1]; \ (o,s,a) \longmapsto \Omega(o,s,a)=p(o|s,a)$ is the observation function.

\end{document}